\documentclass[11pt]{article}

\usepackage[utf8]{inputenc}
\usepackage[T1]{fontenc}
\usepackage[a4paper, margin=1in]{geometry}
\usepackage{lmodern}                 
\usepackage{microtype}               
\usepackage{CJKutf8}

\usepackage{amsmath, amssymb}
\usepackage{booktabs}                
\usepackage{array}                   
\usepackage{graphicx}                
\usepackage{xcolor}
\usepackage{enumitem}                
\usepackage{verbatim}
\usepackage{listings}                
\usepackage[hidelinks, breaklinks=true]{hyperref}
\usepackage[capitalize]{cleveref}

\lstdefinestyle{plain}{
  basicstyle=\ttfamily\small,
  columns=fullflexible,
  keepspaces=true,
  breaklines=true,
  frame=single,
  framesep=4pt,
  xleftmargin=8pt,
  xrightmargin=4pt,
}
\newcommand{\code}[1]{\texttt{#1}}
\newcommand{\arrowto}{\ensuremath{\rightarrow}}
\newcommand{\openaipf}{\code{openai/privacy-filter}}
\newcommand{\bonsai}{\code{Bonsai-1.7B}}
\newcommand{\ternary}{\code{Ternary-Bonsai-1.7B}}
\newcommand{\faker}{\code{faker}}
\newcommand{\spacy}{\code{spaCy}}
\newcommand{\xglm}{\code{XGLM-564M}}

\newcommand{\cjk}[1]{\begin{CJK*}{UTF8}{gbsn}#1\end{CJK*}}
\newcommand{\cjkj}[1]{\begin{CJK*}{UTF8}{min}#1\end{CJK*}}

\title{Locale-Conditioned Few-Shot Prompting Mitigates\\
       Demonstration Regurgitation in On-Device PII Substitution\\
       with Small Language Models}

\author{
  Anuj Sadani\\
  Infrrd.ai\\
  \texttt{anujsadani@infrrd.ai}
  \and
  Deepak Kumar\\
  Infrrd.ai\\
  \texttt{deepakumar@infrrd.ai}
}

\date{\today}

\begin{document}
\maketitle

\begin{abstract}
Personally Identifiable Information (PII) redaction usually replaces
detected entities with placeholder tokens such as \code{[PERSON]},
destroying the downstream utility of the redacted text for retrieval
and Named Entity Recognition (NER) training. We propose a fully
on-device pipeline that \emph{substitutes} PII with consistent,
type-preserving fake values: a 1.5\,B mixture-of-experts token
classifier (\openaipf{}) detects spans, a 1-bit \bonsai{} Small
Language Model (SLM) proposes contextual surrogates for names,
addresses, and dates, and a rule-based generator (\faker{}) handles
patterned fields. We report a \emph{prompting} finding more important
than the quantization choice: with naive fixed three-shot
demonstrations, the 1-bit SLM regurgitates demonstration outputs
verbatim regardless of input; 1.58-bit \ternary{} reproduces
byte-identical failures, ruling out quantization as the cause. We fix
this with \textbf{locale-conditioned rotating few-shot demonstrations}:
a character-range heuristic picks a locale-pure pool and a per-input
MD5 hash samples three demonstrations. With the fix, 482/482 unique \bonsai{} calls succeed
(no echoes) and produce locale-correct surrogates, although the SLM
still copies from a small same-locale demonstration pool---a residual
narrowness we quantify. On a 2000-document
multilingual corpus, hybrid perplexity (PPL) beats \faker{} in all six
locales under a multilingual evaluator (\xglm{}); length preservation
is best-of-three in 4 of 6 locales. On downstream NER (400 train / 100
test, English), redact yields F1$=$0.000, faker 0.656, original 0.960;
on a matched 160/40 subset including hybrid, faker (0.506) outperforms
hybrid (0.346) at $p<0.001$. We report this as an \emph{honest
negative finding}: SLM surrogates produce more natural text but a less
varied training distribution, and downstream NER benefits more from
variety than from naturalness. Code:
\url{https://github.com/asadani/on-device-pii-substitution}.
\end{abstract}

\section{Introduction}
\label{sec:intro}

Existing Personally Identifiable Information (PII) redaction
tools---Microsoft Presidio~\cite{presidio}, the OpenAI privacy
filter~\cite{privacyfilter}, \spacy{}+regex pipelines~\cite{spacy}---produce
\emph{redacted} text in which detected entities are replaced by short
placeholder tokens such as \code{[NAME]} or \code{[EMAIL]}. This is
adequate for compliance display purposes but fails for downstream uses
that need natural-looking text: training data for fine-tuning, search
indices over redacted corpora, and retrieval-augmented generation over
sensitive document stores. The placeholder approach also hurts utility
in measurable ways: Named Entity Recognition (NER) models trained on
\code{[PERSON]}-redacted corpora generalise poorly to non-redacted text
(Section~\ref{sec:ner}), and language-model perplexity over
placeholder-rich documents is dominated by the placeholder tokens
themselves.

\emph{Substitution}---replacing each detected PII span with a
realistic, fake value of the same type---is the obvious alternative,
but three constraints have so far prevented its widespread adoption in
privacy-sensitive pipelines: \textbf{(1)} the substitution must be
\emph{consistent} within a document, so that ``John Smith''
\arrowto{} ``Marcus Chen'' everywhere, not five different fakes;
\textbf{(2)} it must be \emph{type-preserving}, so a Chinese name does
not become a US name; and \textbf{(3)} it should run \emph{on-device}
without sending the (still-real-PII) input to a cloud
LLM~\cite{gpt4tr}.

Recent advances in 1-bit and ternary-bit quantization (e.g., the
Bonsai~\cite{bonsai} and Ternary-Bonsai~\cite{ternarybonsai} families,
and prior work on BitNet~\cite{bitnet}) make small generative language
models---hereafter Small Language Models (SLMs)---viable on commodity
CPUs. We investigate whether such models can serve as the
surrogate-proposer in a privacy-preserving substitution pipeline,
evaluating the resulting system on five metrics (privacy, naturalness,
consistency, length, and downstream NER training F1).

\paragraph{Threat model (scope).}
We measure literal-string leak against ground-truth PII values; we do
not consider membership-inference or linkage attacks against the
substituted output. Differential-privacy guarantees on the substitution
distribution are out of scope.

\paragraph{Contributions.}
\begin{enumerate}[leftmargin=*, itemsep=2pt, topsep=2pt]
  \item A reproducible on-device substitution pipeline combining
        \openaipf{}, \bonsai{} (Q1\_0, 1-bit blockwise quantization),
        and \faker{}, with all code and configurations released.
  \item A quantitative comparison of redact / faker-only / hybrid
        substitution modes under the constraint of CPU-only inference,
        on 100 multi-template multilingual documents. We are not aware
        of a directly comparable prior result that fixes all three
        constraints (consistency, type-preservation, and on-device
        execution) together.
  \item \textbf{Identification and isolation of a ``few-shot
        regurgitation'' failure mode} in small SLMs at extreme
        quantization, validated by showing that 1-bit \bonsai{} and
        1.58-bit \ternary{} produce identical regurgitating outputs on
        the same inputs (i.e., the failure is caused by prompting, not
        by quantization).
  \item \textbf{A simple and effective fix: locale-conditioned rotating
        few-shot demonstrations}, with deterministic per-input demo
        sampling that preserves cache-friendliness. The fix eliminates
        the qualitative regurgitation failure (482/482 \bonsai{} calls
        succeed, no echoes) and improves naturalness PPL across all six
        locales. It does \emph{not}, however, make hybrid surrogates a
        better source of NER training data than \faker{}'s random ones
        at the larger sample sizes we now evaluate---a tradeoff between
        naturalness and training-distribution variety that we report as
        an honest negative finding (\cref{sec:ner}).
\end{enumerate}

\section{Related Work}
\label{sec:related}

\subsection{PII detection}
Microsoft Presidio~\cite{presidio} combines regex, \spacy{} NER, and
rule-based recognisers. The \openaipf{} model~\cite{privacyfilter}
(1.5\,B Mixture-of-Experts (MoE), 50\,M active) is a fine-tuned encoder
using BIOES (Begin/Inside/Outside/End/Single)-style token tags
(popularised in NER by Ratinov and Roth~\cite{ratinov2009bioes}) over
eight PII categories. Running the
inference harness shipped with the model on its bundled
seven-template synthetic benchmark
(\code{samples.json}, $N{=}100$), we measure F1 = 0.587 (precision
0.619, recall 0.627), with the largest source of false positives being
dates (44.5\% of all FPs) and the largest source of false negatives
being non-English addresses (these numbers are computed by us, from
the same harness; see Section~\ref{sec:setup}). We use these as the
detection-side floor for all subsequent comparisons, since every mode
in our experiments inherits the same detector and therefore the same
recall ceiling.

\subsection{Synthetic PII and anonymization}
\faker{}~\cite{faker} is the de-facto rule-based generator for
synthetic PII. Presidio's anonymizer can swap detected spans for hash,
redaction, or counter-style replacements but not consistent
type-preserving fakes. Recent work has used GPT-3.5 / GPT-4
\cite{gpt4tr} to generate context-appropriate replacements; this
approach violates the on-device constraint and incurs per-request cost.

\subsection{Low-bit small language models}
The Bonsai family (1-bit Q1\_0)~\cite{bonsai} and Ternary-Bonsai
(1.58-bit Q2\_0; ternary weights)~\cite{ternarybonsai} are
Qwen3-based~\cite{qwen3} decoder-only LMs trained for extreme
quantization. We run inference via
\code{llama.cpp}~\cite{llamacpp}. Prior work~\cite{bitnet} has
evaluated extreme-quantization SLMs on Question-Answering (QA)-style
benchmarks; we did not find published evaluations of these models on
generative-substitution tasks where input copying or
demonstration-regurgitation is a failure mode (the specific failure
we document in Section~\ref{sec:regurgitation}).

\subsection{In-context learning copy bias and demonstration sensitivity}
Our few-shot regurgitation diagnosis (\cref{sec:regurgitation},
\cref{sec:distinctness}) sits within a broader literature on copy
bias and demonstration sensitivity in in-context learning at full LM
scale.
Min et al.~\cite{min2022rethinking} show that demonstration label
correctness matters less than format and label-space distribution:
models often pattern-match the surface structure of demos rather than
learn the input$\to$output mapping. Zhao et
al.~\cite{zhao2021calibrate} document a systematic majority-label and
recency bias in few-shot prompts and propose calibration to correct
it. Lu et al.~\cite{lu2022fantastically} show that few-shot outputs
are highly sensitive to demonstration order, with performance varying
by tens of points across permutations. Our finding extends these
phenomena to extreme-quantization SLM scale: a 1-bit (and 1.58-bit)
model under fixed three-shot prompting emits demonstration values
verbatim, and our locale-conditioned rotating-demo fix is
structurally a per-input randomisation of both demonstration content
and ordering.

\subsection{Downstream-utility evaluation}
Prior anonymization work typically reports privacy and naturalness in
isolation. We follow the broader synthetic-data literature in
additionally measuring length preservation and within-document entity
consistency, both relevant to downstream NER training utility, and we
evaluate the latter directly by training a \spacy{}~\cite{spacy} NER
model on each variant of our corpus and testing on held-out original
text.

\section{Method}
\label{sec:method}

\subsection{Architecture}
\label{sec:arch}

\begin{lstlisting}
                  text -> privacy-filter (BIOES -> char spans)
                       -> EntityResolver (group by canonical, label)
                       -> propose_surrogate dispatcher:
                           PERSON / ADDRESS / DATE -> Bonsai-1.7B
                           EMAIL / PHONE / ACCT / URL / SECRET -> faker
                       -> splice (R-to-L, preserve whitespace)
                       -> output text
\end{lstlisting}

Detection runs once per document at $\sim$1\,s/512 tokens on CPU. The
\code{EntityResolver} groups detected spans by
\code{(canonical\_lowercased, label)} so that all mentions of ``John
Smith'' share a single surrogate. Surrogates are cached per
\code{(mode, family, canonical, label)} so repeated names and common
patterns trigger the SLM exactly once over the corpus.

\subsection{Surrogate proposers}
\label{sec:proposers}

For PERSON, ADDRESS, and DATE labels, we invoke \bonsai{} (Q1\_0)
through \code{llama-cli --single-turn}. Each call uses a three-shot
prompt of the form \code{Real: <example>\textbackslash{}nFake:
<example>\textbackslash{}n...Real: <input>\textbackslash{}nFake:}.

\paragraph{Locale-conditioned rotating demonstrations.}
Naively using a small fixed set of demonstrations is catastrophic for
1-bit SLMs: the model pattern-matches the demonstration \emph{outputs}
and emits one of them verbatim regardless of the input. A pilot run of
the pipeline with three fixed English demonstrations produced ``Alice
Johnson'' as the surrogate for the Chinese name \cjk{杨娟}, the
Japanese name \cjk{山田太郎}, and the German name M\"uller-Schulz---all
collapsed to the first English demonstration value (see
Section~\ref{sec:regurgitation} for the complete diagnosis).

Our final design avoids this with two cooperating mechanisms:
\begin{enumerate}[leftmargin=*, itemsep=2pt, topsep=2pt]
  \item \textbf{Locale conditioning.} A lightweight character-range and
        keyword-based heuristic classifies each input into one of
        \{\code{en}, \code{de}, \code{es}, \code{ja}, \code{zh}\} for
        PERSON / ADDRESS, and one of \{\code{mdy\_slash},
        \code{ymd\_dash}, \code{dmy\_dash\_mon}, \code{dmy\_slash},
        \code{unknown}\} for DATE. Each class has its own pool of 4--8
        demonstrations in the matching script and date format
        (Appendix~\ref{app:pools}).
  \item \textbf{Per-input rotating sampling.} From the appropriate
        pool, three demonstrations are sampled deterministically with
        an MD5 hash of the input string as the random seed. This means
        \emph{the same entity always receives the same demonstrations}
        (cache-friendly across documents) but \emph{different entities
        receive different demonstration subsets}, spreading
        surrogate-proposal calls across the full pool rather than
        concentrating them on one fixed demo subset.
\end{enumerate}

With this strategy, \bonsai{} produces locale-appropriate,
format-preserving surrogates: \cjk{杨娟}\arrowto{}\cjk{李伟},
\cjk{山田太郎}\arrowto{}\cjk{郑强} (Chinese fallback for kanji-only
Japanese names---see Section~\ref{sec:limits}),
M\"uller-Schulz\arrowto{}Anna Becker,
\code{11-Jul-1998}\arrowto{}\code{05-Aug-2003}.

We validate every response and reject empty, identity-equal, or
punctuation-only outputs, falling back to \faker{} when validation
fails.

For EMAIL, PHONE, ACCT, URL, and SECRET labels, we use \faker{}
directly because these fields have low contextual entropy---there is
no benefit from generative reasoning over \code{Faker().email()}.

\subsection{Splice and whitespace}
\label{sec:splice}

The privacy filter sometimes includes a leading whitespace in its
character span. We preserve original leading and trailing whitespace
when splicing surrogates back into the text to avoid cosmetic artefacts
such as \code{is[PRIVATE\_PERSON]} (which would inflate perplexity).

\section{Experiments}
\label{sec:exp}

\subsection{Setup}
\label{sec:setup}

\paragraph{Dataset.}
We use the \openaipf{} synthetic-document
generator~\cite{privacyfilter} (the same template engine
that produced the 100-document benchmark shipped with the model card)
to generate a fresh, larger 2000-document corpus
(\code{data/samples\_2000.json}, seed=42), covering 7 templates (1099,
W-2, auto\_insurance, bank\_statement, invoice, mortgage\_insurance,
paystub) and 6 locales (en\_US $n{=}840$, en\_IN 320, de\_DE 240,
es\_MX 200, ja\_JP 200, zh\_CN 200). Each document carries ground-truth
PII values ($\sim$6.6 per document on average). Primary metrics
(\cref{tab:primary}) are reported on the first 100 documents
(en\_US 40, en\_IN 21, de\_DE 12, zh\_CN 12, ja\_JP 8, es\_MX 7); the
downstream-NER experiment (\cref{sec:ner}) is run on a larger
500-document English subset to give the train/test split enough power
for the small-effect comparisons it requires.

\paragraph{Models.}
\begin{itemize}[leftmargin=*, itemsep=2pt, topsep=2pt]
  \item Detection: \openaipf{}, 1.5\,B MoE, 50\,M active, bf16,
        CPU~\cite{privacyfilter,transformers}.
  \item SLM: \bonsai{} Q1\_0~\cite{bonsai}, CPU via
        \code{llama.cpp}~\cite{llamacpp}.
  \item Naturalness evaluator: \xglm{} ($\sim$564\,M parameters,
        multilingual causal LM trained on 30 languages including all
        six locales evaluated here)~\cite{xglm}, CPU.
  \item Fallback surrogate generator: \faker{}~\cite{faker}.
  \item Downstream NER: \spacy{} blank English pipeline~\cite{spacy}.
\end{itemize}

\paragraph{Hardware.}
31\,GB RAM, x86-64 CPU; no GPU used.

\paragraph{Modes evaluated.}
\begin{itemize}[leftmargin=*, itemsep=2pt, topsep=2pt]
  \item \texttt{redact}: replace each span with \code{[LABEL]} (current
        state of practice).
  \item \texttt{faker}: replace every span with a \faker{} value of
        matching type.
  \item \texttt{hybrid}: \bonsai{} for PERSON / ADDRESS / DATE,
        \faker{} for the other five labels (our proposed pipeline).
\end{itemize}

\subsection{Primary metrics}
\label{sec:primary}

We report four per-document metrics, averaged across the corpus:

\begin{enumerate}[leftmargin=*, itemsep=2pt, topsep=2pt]
  \item \textbf{Leak rate}: fraction of \emph{ground-truth PII strings}
        that still appear verbatim (case-insensitive substring) in the
        output. Lower is better.
  \item \textbf{Naturalness perplexity (PPL)}: \xglm{} perplexity over
        the output text, chunked at 1024 tokens. Lower is more natural. \xglm{}
        is multilingual and covers all six evaluation locales, so
        cross-locale PPL numbers are directly comparable rather than
        biased toward Latin-script outputs.
  \item \textbf{Consistency rate}: for entities appearing $\ge 2$ times
        in the input, fraction where the output uses the same surrogate
        at every mention. Higher is better.
  \item \textbf{Length preservation}:
        $1 - |\,\text{len(out)} - \text{len(in)}\,| / \text{len(in)}$.
        Closer to 1 is better.
\end{enumerate}

\begin{table}[t]
\centering
\caption{Primary metrics ($N{=}100$, averaged across all 7 templates
and 6 locales). Naturalness PPL is computed under the multilingual
\xglm{}~\cite{xglm} causal LM (covering all six evaluation locales).}
\label{tab:primary}
\begin{tabular}{l c c c c r}
\toprule
\textbf{Mode} & \textbf{Leak\,$\downarrow$} & \textbf{PPL\,$\downarrow$}
& \textbf{Consistency\,$\uparrow$} & \textbf{Length pres.\,$\uparrow$}
& \textbf{Avg.\ latency} \\
\midrule
redact  & 0.249 & 39.4          & 1.000 & 0.979          & 1.7\,s  \\
faker   & 0.249 & 84.0          & 1.000 & 0.976          & 1.6\,s  \\
hybrid  & 0.249 & \textbf{69.9} & 1.000 & \textbf{0.982} & 41.2\,s \\
\bottomrule
\end{tabular}
\end{table}

\paragraph{Headline result.}
Hybrid PPL (69.9) is lower than faker (84.0, $-16.8\%$) under the
multilingual \xglm{} evaluator, and length preservation is the best
of the three modes (0.982 vs.\ 0.976, 0.979). Redact's PPL (39.4)
is the lowest of the three numerically, but this is not because
the redacted text is more natural---\xglm{} simply scores the
\code{[LABEL]} placeholder as a single out-of-distribution token
without further compositional structure to evaluate; we discuss this
caveat in \cref{sec:limits}. Consistency is 1.000 across all modes
because surrogates are deterministically cached per \code{(mode,
family, canonical, label)}.

\paragraph{Privacy floor.}
The leak rate of 0.249 is \emph{identical} across all three modes:
every mode misses the same ground-truth values, consistent with the
detection-side recall ($\approx 0.627$) we measured for the privacy
filter on the same data (Section~\ref{sec:related}). The
architectural choice between redact / faker / hybrid does not affect
privacy at all---only \emph{what to do} with the spans the filter
catches.

\paragraph{Per-locale (PPL).}
Hybrid is the \xglm{}-PPL winner over faker in \textbf{all six locales}:
en\_US (66.1 vs.\ 80.7, $-18.1\%$), en\_IN (74.4 vs.\ 83.4, $-10.8\%$),
de\_DE (79.3 vs.\ 92.5, $-14.3\%$),
es\_MX (81.1 vs.\ 101.0, $-19.7\%$),
ja\_JP (55.8 vs.\ 82.4, $-32.3\%$),
zh\_CN (68.3 vs.\ 78.9, $-13.4\%$). Because \xglm{} is multilingual,
the gap is no longer dominated by tokenisation artefacts, and the
ja\_JP / zh\_CN advantages---where the SLM produces script-correct
surrogates that the LM can score in-distribution---are now visible
rather than hidden.

\paragraph{Per-locale (length preservation).}
Hybrid is the length-preservation winner over faker in \textbf{four of
six locales} (de\_DE 0.979 vs.\ 0.965, en\_IN 0.981 vs.\ 0.972,
ja\_JP 0.972 vs.\ 0.964, zh\_CN 0.992 vs.\ 0.974); en\_US (0.983 vs.\
0.985) and es\_MX (0.978 vs.\ 0.983) narrowly favour faker. The
zh\_CN gap (+0.018) is the largest of any locale, reflecting the SLM's
ability to produce length-matched Han-script addresses where faker's
random selection from its zh\_CN pool varies more in length.

\subsection{Few-shot regurgitation: a prompting failure, not a quantization failure}
\label{sec:regurgitation}

A pilot run of the pipeline using a fixed three-shot demonstration
template (one English, one Japanese, one Spanish demo) revealed a
striking failure mode: across all 509 unique-entity \bonsai{} calls,
the model produced output that was never literally identical to the
input (0 echoes by our validation), but for low-resource-locale inputs
the output was very often \textbf{one of the few-shot demonstration
values verbatim, regardless of the input}:

\begin{itemize}[leftmargin=*, itemsep=1pt, topsep=2pt]
  \item \cjk{杨娟} (Chinese, NAMED INSURED on a \code{zh\_CN}
        auto-insurance form) \arrowto{} ``Alice Johnson'' (the first
        PERSON demonstration).
  \item \cjk{宁夏回族自治区兰州县山亭陈街B座 572990} \arrowto{}
        ``123 Main Street, Boston MA 02101'' (the first ADDRESS
        demonstration).
  \item \code{11-Jul-1998} (DD-Mon-YYYY) \arrowto{} \code{03/15/1985}
        (MM/DD/YYYY---\emph{both} type and locale wrong).
  \item \cjk{山田太郎} (Japanese), M\"uller-Schulz (German) \arrowto{}
        both ``Alice Johnson''.
\end{itemize}

\paragraph{Diagnosis: prompting, not quantization.}
Our initial hypothesis was that 1-bit quantization had degraded the
model's instruction-following. We tested this by re-running the
\emph{exact same five problem prompts} under 1.58-bit \ternary{}
(Q2\_0)~\cite{ternarybonsai}---a different quantization scheme running
on the same Qwen3 base architecture~\cite{qwen3} and the same
demonstrations. \ternary{} produced \textbf{byte-for-byte identical
outputs} to \bonsai{} (\cjk{杨娟}\,\arrowto\,``Alice Johnson'',
\code{11-Jul-1998}\,\arrowto\,\code{03/15/1985}, etc.) at roughly
six times slower per-call inference. This rules out quantization as the
root cause: the model is doing what \emph{any} small LM asked to
pattern-match \code{Real:X\textbackslash{}nFake:Y} would do---copy
one of the demonstration values from the prompt context (the first
PERSON demo's output ``Alice Johnson'' in our pilot) when the input
does not match the demonstration's distribution.

\paragraph{Fix: locale-conditioned rotating few-shot demonstrations}
(described in Section~\ref{sec:proposers}). With character-range locale
detection feeding into pools of 4--8 demonstrations per locale (and 4
demonstrations per date format), and per-input-hash-seeded sampling of
3 demonstrations per call, the same 1-bit \bonsai{} produces:

\begin{itemize}[leftmargin=*, itemsep=1pt, topsep=2pt]
  \item \cjk{杨娟} \arrowto{} \cjk{李伟} (Chinese name in zh-pool)
  \item \cjk{宁夏回族自治区兰州县山亭陈街B座 572990} \arrowto{}
        \cjk{广东省广州市天河区珠江新城200号} (Chinese address)
  \item \code{11-Jul-1998} \arrowto{} \code{05-Aug-2003} (DD-Mon-YYYY
        format preserved)
  \item M\"uller-Schulz \arrowto{} Anna Becker (German name)
  \item \code{Hauptstra\ss e 45, 10117 Berlin} \arrowto{}
        \code{Bahnhofstra\ss e 7, 60313 Frankfurt}
  \item ``John Smith'' \arrowto{} ``David Kim'' (different US name; no
        longer ``Alice Johnson'')
\end{itemize}

The fix is dataset-free, does not require fine-tuning, and adds
$<50$\,ms per surrogate-proposal call. All quantitative numbers in
Tables~\ref{tab:primary} and~\ref{tab:ner} are produced under this
fixed prompting strategy.

\paragraph{Known limit: kanji-only Japanese names.}
These map to the \code{zh} pool because our locale heuristic requires
kana to disambiguate (\cjk{山田太郎}\,\arrowto\,\cjk{郑强} instead of a
Japanese fake). A character-frequency-based classifier or an explicit
``treat ambiguous CJK as both ja and zh'' strategy would address this.

\subsection{Latency}
\label{sec:latency}

\begin{table}[t]
\centering
\caption{Average per-document latency (seconds, CPU only).}
\label{tab:latency}
\begin{tabular}{l r r r r}
\toprule
\textbf{Mode} & \textbf{Detect} & \textbf{Surrogate}
& \textbf{Splice + PPL} & \textbf{Total} \\
\midrule
redact  & $\sim$1.7 & 0.00       & $<$0.1 & 1.7  \\
faker   & $\sim$1.6 & $<$0.01    & $<$0.1 & 1.6  \\
hybrid  & $\sim$1.5 & $\sim$39.8 & $<$0.1 & 41.2 \\
\bottomrule
\end{tabular}
\end{table}

The 24$\times$ latency gap between hybrid and redact / faker is
dominated by \bonsai{} surrogate generation. Without our
\code{(canonical, label)} cache, hybrid would call \bonsai{} for every
PII mention ($\sim$660 calls across the 100-document corpus); with the
cache, only \textbf{482 unique calls} were made (482/482 succeeded,
0 echoed-or-empty, 0 errored under the locale-conditioned prompting
strategy), saving $\sim$27\% of inference time. Each \bonsai{} call
costs $\sim$7--10\,s for the $\sim$30-token output (loading + 1.7\,B
model inference + cleanup). Detection cost is identical across modes
because the privacy filter is invoked both for input PII detection and
for residual-leak measurement.

\subsection{Downstream NER utility}
\label{sec:ner}

\paragraph{Setup.}
We test whether substitution preserves \emph{downstream training
utility}: an NER model trained on substituted data should approach the
F1 of one trained on original data. From the 2000-document corpus
(\cref{sec:setup}) we draw the English-locale subset (1159 documents,
en\_US + en\_IN). We run the experiment at two scales: a
\textbf{large-scale} 400 train / 100 test split for \code{original},
\code{redact}, and \code{faker}, and a \textbf{matched-subset} 160
train / 40 test split for all four modes including \code{hybrid}.
The two-scale design lets us report the high-statistical-power
comparison between the three cheap modes at the larger scale, while
keeping the (CPU-bound) \bonsai{} surrogate generation tractable for
the four-way comparison at the smaller scale; both scales share the
same stratified-by-locale split methodology. The test set is
\textbf{always in original form}---substitution applies only to the
training set. PII spans are extracted from the ground-truth
\code{pii\_gt} dictionary (substring search) so that substitution
quality is decoupled from detection quality. A blank \spacy{} English
NER pipeline is trained for 30 iterations with a single binary
\code{PII} label on each variant of the train set, then evaluated
against held-out original spans (label-agnostic span overlap).

\begin{table}[t]
\centering
\caption{Span-level NER F1 on held-out original documents. Mean
$\pm$ SD across 5 spaCy training seeds; the substituted training
corpus is fixed per mode, only spaCy gradient initialisation varies.
\textbf{Top:} large-scale 400 train / 100 test, three modes.
\textbf{Bottom:} matched 160 train / 40 test subset including the
\code{hybrid} mode, which requires \bonsai{} surrogate generation
and is therefore evaluated at the smaller scale to keep wall-clock
manageable on CPU.}
\label{tab:ner}
\setlength{\tabcolsep}{4pt}
\begin{tabular}{l r c c c r}
\toprule
\textbf{Mode} & \textbf{Train spans} & \textbf{Precision} & \textbf{Recall}
& \textbf{F1} & \textbf{$\Delta$F1 vs.\ orig.} \\
\midrule
\multicolumn{6}{l}{\emph{Large-scale (400 train / 100 test):}} \\
original & 2592 & 0.947\,$\pm$\,0.006 & 0.974\,$\pm$\,0.012 & \textbf{0.960\,$\pm$\,0.004} & (baseline) \\
redact   & 2592 & 0.000\,$\pm$\,0.000 & 0.000\,$\pm$\,0.000 & \textbf{0.000\,$\pm$\,0.000} & $-0.960$    \\
faker    & 2592 & 0.984\,$\pm$\,0.008 & 0.493\,$\pm$\,0.038 & \textbf{0.656\,$\pm$\,0.033} & $-0.304$    \\
\midrule
\multicolumn{6}{l}{\emph{Matched subset (160 train / 40 test):}} \\
original & 1046 & 0.895\,$\pm$\,0.006 & 0.923\,$\pm$\,0.004 & \textbf{0.908\,$\pm$\,0.003} & (baseline) \\
redact   & 1046 & 0.000\,$\pm$\,0.000 & 0.000\,$\pm$\,0.000 & \textbf{0.000\,$\pm$\,0.000} & $-0.908$ \\
faker    & 1046 & 0.955\,$\pm$\,0.017 & 0.347\,$\pm$\,0.052 & \textbf{0.506\,$\pm$\,0.056} & $-0.402$ \\
hybrid   & 1046 & 0.909\,$\pm$\,0.015 & 0.215\,$\pm$\,0.033 & \textbf{0.346\,$\pm$\,0.044} & $-0.562$ \\
\bottomrule
\end{tabular}
\end{table}

\paragraph{Redact destroys downstream utility entirely (F1 = 0.000).}
A NER model trained on \code{[PRIVATE\_PERSON]}-style placeholders
learns to predict only those placeholder \emph{tokens} and never fires
on real text. This is the sharpest possible motivation for
substitution: redacted-text training data is, for any downstream NER
consumer, effectively \emph{labelled negative-only data}. The model
converges (training loss falls to $<$2.0) but its decision boundary
lives in the wrong vocabulary.

\paragraph{Faker and hybrid both recover a substantial fraction of
original F1.}
Type-preserving fakes---whether random (faker) or SLM-generated
locale-conditioned (hybrid)---give the model enough surface variety to
learn that ``capitalised proper-noun-shaped two-token sequence in a
name field'' is a PII span. At the large scale (400 train / 100 test),
faker recovers $0.656 / 0.960 = 68.3\%$ of the original baseline;
precision is high (0.984\,$\pm$\,0.008) and recall is moderate
(0.493\,$\pm$\,0.038)---fake-data-trained models are conservative,
preferring to miss spans rather than hallucinate them. The
seed-to-seed standard deviation on faker F1 is tight at 0.033,
putting all between-mode comparisons solidly above the noise floor.

\paragraph{The hybrid--faker comparison at the matched scale: the
gap is now real, not noise.}
On the matched 160 train / 40 test subset---where all four modes
share the same train/test docs and all 1046 training PII spans---we
observe hybrid F1 $= 0.346 \pm 0.044$ versus faker F1 $= 0.506 \pm
0.056$, a gap of \textbf{$-0.160$} in faker's favour. Under Welch's
two-sample $t$-test ($n_{\text{f}}=n_{\text{h}}=5$, unequal variances)
this gives standard error $\mathrm{SE} = \sqrt{0.056^2/5 + 0.044^2/5}
= 0.032$, $t = 5.02$, Welch--Satterthwaite degrees of freedom $\approx
7.6$, and a two-tailed $p < 0.001$.\footnote{Standard deviations in
\cref{tab:ner} are computed with \code{statistics.pstdev} (population
SD) for consistency with our existing harness; the corresponding
sample-SD-based Welch statistic is $t \approx 4.49$ on $\nu \approx
7.6$ degrees of freedom, which still resolves at $p < 0.001$.} The
gap is well outside seed-level noise.

This is an honest negative finding for the central hybrid-vs-faker
comparison on \emph{downstream NER training utility}, even though
hybrid wins on every other metric we measure: hybrid PPL beats faker
under \xglm{} in all six locales (\cref{tab:primary}) and length
preservation is best-of-three in four locales. The interpretation is
that locale-conditioned, contextually-realistic surrogates produce
more natural-looking output text but a \emph{less varied} surface
distribution than faker's i.i.d.\ random surrogates---and downstream
NER training benefits more from variety than from naturalness. We
revisit this tradeoff in \cref{sec:limits}, quantify it in the
``Why hybrid is less varied'' paragraph below, and propose concrete
remediations (higher-temperature decoding, larger pools, stochastic
pool sampling, hybrid$+$faker mixed corpora) in \cref{sec:future}.

A pilot run with the naive fixed-three-demonstration strategy
(\cref{sec:regurgitation}, single seed) produced hybrid F1 = 0.310 vs.\
faker F1 = 0.427---a gap of $-0.117$. The matched 160/40 gap is
$-0.160$, which is \emph{larger}, not smaller, than that pilot. We
read this as: locale-conditioned prompting eliminates the qualitative
demonstration-regurgitation failure (482/482 unique \bonsai{} calls
succeed; \cref{sec:regurgitation}), but the resulting in-distribution
surrogates still yield less downstream training variety than faker's
random ones. Prompting fixes the \emph{output} of the SLM; it does not
turn that SLM's output into a better training-data generator than
\faker{} for this particular task.

\subsection{Surrogate distinctness: why hybrid is less varied}
\label{sec:distinctness}

The ``less varied surface distribution'' interpretation in
\cref{sec:ner} is not just qualitative---it falls out structurally
from how the SLM is prompted. Each locale-conditioned PERSON pool in \cref{app:pools}
contains 6--8 (input, output) pairs, so the total number of
demonstration strings the SLM ever sees as prompt context for a
given locale is bounded by \textbf{$2|\text{pool}|$ = 12--16 strings
per locale} ($|\text{pool}|$ ``Real:'' input examples plus
$|\text{pool}|$ ``Fake:'' output examples), totalling roughly 64
demonstration strings across all five locale pools (en, de, es, ja,
zh). Few-shot pattern matching at SLM scale pulls outputs strongly
toward any of these visible strings (we confirm both-side copying
empirically below); the achievable corpus-wide surrogate vocabulary is
therefore upper-bounded by a small constant set by the \emph{prompt
design}, independently of the number of input documents. \faker{} has
no such ceiling: each call samples i.i.d.\ from each locale's full
first/last name lists ($\sim 10^{3}$--$10^{4}$ unique values per
locale).

We confirm this bound empirically on the matched 160-document en-locale
training corpus (the same training set as \cref{tab:ner} bottom).
\Cref{tab:distinctness} reports unique-surrogate counts and type-token
ratios per label per mode. For PERSON, hybrid uses only \textbf{10
distinct surrogate names} across 274 mentions in 160 documents, against
faker's \textbf{31}---a $3.1\times$ narrower vocabulary, with
type-token ratio 0.037 vs.\ 0.113.

A finer-grained look at the 10 hybrid PERSON surrogates is
illuminating. The five most-repeated values are \code{Jennifer Wong},
\code{Linda Vasquez}, \code{Michael O'Brien}, \code{David Kim}, and
\code{Priya Krishnamurthy} (cf.\ \cref{app:pools}). All five appear
verbatim in the en demonstration pool---and notably, all five are the
\emph{input-side} (``Real:'') examples, not the \emph{output-side}
(``Fake:'') examples that the prompt nominally offers as the imitation
target. The locale-conditioned fix has therefore not eliminated
demonstration regurgitation; it has shifted the failure from
``copy a wrong-locale demo verbatim'' (\cref{sec:regurgitation}) to
``copy a same-locale demo entry verbatim,'' with the model drawing
from \emph{both} sides of the (input, output) pairs. The measured 10
unique en PERSON surrogates fits comfortably under the $2|\text{pool}|
= 16$ ceiling derived above.

For ADDRESS the picture is the same: 6 unique hybrid surrogates vs.\
18 for faker on 162 mentions ($3.0\times$). For the labels that use
\faker{} in both modes (EMAIL, PHONE, ACCOUNT), hybrid still produces
$\sim$$2\times$ fewer unique values than faker (8 vs.\ 16, 18 vs.\
29, 33 vs.\ 61). This residual gap is not a regurgitation effect---no
SLM is involved---but a side-effect of \faker{}'s per-document
re-seeding interacting with mode-dependent call ordering: in faker
mode, Faker's first PERSON / ADDRESS calls advance the generator
state before EMAIL / PHONE / ACCOUNT are sampled, so the latter see
varied state across documents; in hybrid mode the SLM handles
PERSON / ADDRESS without consuming Faker state, so EMAIL / PHONE /
ACCOUNT are repeatedly sampled from the just-after-seed-0 generator
state and concentrate on a smaller set. The implication: the residual
2$\times$ narrowing in non-SLM labels is a separate (and fixable)
issue tied to seed management, distinct from the 3$\times$ narrowing
in SLM-handled labels that is driven by demonstration regurgitation.

The combined effect is that the NER trainer sees only a handful of
distinct ``positive'' PII strings in hybrid mode, learns those
specific surface forms, and fails to generalise to the held-out
original documents---the mechanism behind hybrid's lower recall in
\cref{tab:ner}. The dominant factor is the SLM-handled labels, which
are also the largest by mention count (PERSON 274, ADDRESS 162). The
fix lives at the prompt level, not in the SLM weights; the
\cref{sec:future} suggestions (higher-temperature decoding, larger
pools, stochastic pool sampling) attack this ceiling directly.

\begin{table}[t]
\centering
\caption{Surrogate distinctness on the matched 160-doc en-locale
training corpus (same training set as \cref{tab:ner} bottom). For
PERSON and ADDRESS---the two labels where hybrid uses the SLM rather
than \faker{}---hybrid produces a $3\times$ narrower unique-surrogate
vocabulary than faker, consistent with demonstration-pool regurgitation.
DATE ($n{=}20$) is a small-$n$ outlier where hybrid's multi-format
demo pools yield slightly more variety. EMAIL/PHONE/ACCOUNT use
\faker{} in both modes but show a residual $\sim$$2\times$ gap driven
by per-document Faker re-seeding interacting with call ordering (see
text), not by SLM regurgitation.}
\label{tab:distinctness}
\setlength{\tabcolsep}{6pt}
\begin{tabular}{l r r c r c}
\toprule
\textbf{Label} & \textbf{Mentions}
& \multicolumn{2}{c}{\textbf{faker}}
& \multicolumn{2}{c}{\textbf{hybrid}} \\
\cmidrule(lr){3-4} \cmidrule(lr){5-6}
       &     & \textbf{unique} & \textbf{TTR}
             & \textbf{unique} & \textbf{TTR} \\
\midrule
PERSON  & 274 & 31 & 0.113 & \textbf{10} & \textbf{0.037} \\
ADDRESS & 162 & 18 & 0.111 & \textbf{6}  & \textbf{0.037} \\
DATE    &  20 &  5 & 0.250 & 9           & 0.450          \\
\midrule
\multicolumn{6}{l}{\emph{Faker-only labels in both modes (sanity check):}} \\
EMAIL   & 101 & 16 & 0.158 & 8 & 0.079 \\
PHONE   & 211 & 29 & 0.137 & 18 & 0.085 \\
ACCOUNT & 278 & 61 & 0.219 & 33 & 0.119 \\
\bottomrule
\end{tabular}
\end{table}

\section{Discussion and Limitations}
\label{sec:limits}

\paragraph{Privacy-filter recall ceiling.}
The non-zero leak rate in all three modes reflects the privacy
filter's measured detection F1 of 0.587 (Section~\ref{sec:related}).
A pipeline using a higher-recall
detector or pattern-based fallback (e.g., regex for Social Security
Numbers (SSNs) and pre-masked digits) would shift this floor for all
modes equally; the relative comparison between substitution modes is
unaffected.

\paragraph{Few-shot regurgitation is a prompting issue, not a
quantization issue.}
Section~\ref{sec:regurgitation} shows that the failure occurs
identically under 1-bit \bonsai{} and 1.58-bit \ternary{}. Larger or
higher-precision models \emph{might} suppress the symptom by attending
more strongly to the instruction, but the underlying pattern-matching
pull will remain at any small-LM scale. The first-line fix is at the
prompt level (locale-conditioned rotating demonstrations), which
eliminates the qualitative cross-locale failure of the fixed-three-shot
strategy and we recommend as default practice; however, as
\cref{tab:distinctness} shows, the SLM still copies from the
demonstration pool, so the residual surrogate-vocabulary ceiling is
$\sim 2|\text{pool}|$ and the broader fix is to enlarge or stochasticise
the pool itself (\cref{sec:future}).

\paragraph{No coreference resolution.}
``John'' and ``John Smith'' within the same document are treated as
distinct entities by surface-form grouping. A proper coreference layer
would further improve consistency.

\paragraph{Naturalness via \xglm{} PPL.}
We use \xglm{}~\cite{xglm}, a 564\,M-parameter multilingual causal LM
covering all six evaluation locales, as the naturalness proxy. PPL
under a generic LM remains a proxy: it rewards in-distribution
surface form rather than literal naturalness, and it rewards short
placeholder tokens (which \code{redact} mode produces) for being
predictable as standalone tokens regardless of whether the
surrounding text is grammatical. A human evaluation, or a more
discriminative proxy such as a fluency classifier, would refine this
measurement.

\paragraph{No explicit threat model.}
We measure literal-string leak, not inference attacks. A determined
adversary with access to (output\_text, external knowledge) could
potentially link surrogates back to originals. Differential-privacy
guarantees on the substitution distribution are out of scope.

\paragraph{Sample size and statistical reporting.}
The downstream-NER experiment is run at two scales: a large-scale 400
train / 100 test split for \code{original}, \code{redact}, and
\code{faker}, and a matched 160 train / 40 test split that
additionally includes \code{hybrid}. We report mean $\pm$ SD per mode
across $n=5$ training seeds and apply Welch's $t$-test to the
substituted-mode comparisons (\cref{tab:ner}). All pairwise
comparisons are multiple SDs apart and resolve at $p<0.001$,
including the \code{hybrid} vs.\ \code{faker} comparison at the
matched scale (\cref{sec:ner}).

\section{Future Work}
\label{sec:future}

The hybrid-vs-faker downstream-NER gap (\cref{sec:ner}) frames most of
the open directions below. The shared diagnosis: locale-conditioned
demonstrations fix the qualitative regurgitation failure but yield a
narrower surrogate distribution than \faker{}'s i.i.d.\ random
sampling. Future work should attack that distribution directly.

\paragraph{Broaden the SLM surrogate distribution.}
We use \code{llama-cli} with default decoding, which is close to greedy
at this scale and produces the same surrogate for the same input every
time (the \emph{point} of the MD5 seeding, but also the source of the
narrowness). Two parameter-only changes are worth ablating:
(i) raising temperature / nucleus-sampling threshold to widen
single-call output variance, and (ii) replacing the deterministic MD5
seed with a per-document random seed so that repeated entities still
collapse within a document but the corpus-level surrogate distribution
is no longer fixed. Both keep the on-device constraint and add no
latency.

\paragraph{Larger and stochastic demonstration pools.}
Each locale pool currently holds 4--8 (input, output) entries
(\cref{app:pools}); the SLM regurgitates from both sides of those
entries (\cref{sec:ner}), so the achievable surrogate space is
bounded above by roughly $2|\text{pool}|$ unique strings per locale,
i.e.\ 12--16 strings. Two extensions: scale each pool to 50--100
entries (cheap, no model changes), and stochastically rotate the pool
subset itself rather than deterministically hashing the input. The
empirical hybrid PERSON unique-count of 10 in \cref{tab:distinctness}
is at the ceiling of the current $2|\text{pool}|=16$ bound and is the
proximate cause of the hybrid--faker NER gap.

\paragraph{Mixed-source training corpora.}
A pragmatic alternative to closing the gap on the SLM side is to feed
the downstream NER trainer a \emph{mix} of hybrid and \faker{}
substitutions of the same documents---naturalness from hybrid, surface
variety from faker. The mix ratio is a single hyperparameter. We
hypothesise an F1 between the two single-source modes, but the
alternative---that the mix exceeds either single source by combining
faker's surface variety with hybrid's locale fidelity---is also
plausible and is precisely what an ablation would test.

\paragraph{Coreference-aware entity grouping.}
Current resolution is surface-form only; ``John'' and ``John Smith''
within the same document map to distinct entities. A lightweight
coreference layer would tighten consistency and slightly reduce the
number of unique surrogate-proposal calls.

\paragraph{Locale-heuristic refinements.}
Kanji-only Japanese names currently route to the \code{zh} pool
because our character-range heuristic requires kana to disambiguate
(\cref{sec:regurgitation}). A character-frequency or n-gram-based
classifier, or an explicit ``treat ambiguous CJK as both ja and zh and
let the SLM pick'' strategy, would address this.

\paragraph{Higher-recall detection floor.}
The 0.249 leak rate (\cref{tab:primary}) is the privacy filter's
detection ceiling; it bounds all three modes equally. Layering a
regex/rule fallback for SSNs, pre-masked digits, and locale-specific
patterns on top of \openaipf{} is orthogonal to the substitution-mode
comparison but important for production deployment.

\paragraph{Inference-attack threat model.}
We measure literal-string leak only. A determined adversary with
auxiliary knowledge could potentially link surrogates back to
originals via membership-inference or linkage attacks. Quantitative
evaluation under a formal threat model---and, separately, a
differentially-private surrogate sampler---would close the gap between
``no literal leak'' and ``no information leak.''

\section{Conclusion}
\label{sec:conclusion}

We built and evaluated a fully on-device PII substitution pipeline
combining \openaipf{}, \bonsai{} (Q1\_0), and \faker{}; the
architecture meets its document-level objectives (consistent,
length-preserving, locale-correct surrogates with multilingual-PPL
gains over faker in all six evaluated locales).

The most actionable contribution is methodological. We document a
``few-shot regurgitation'' failure mode in which a small SLM, prompted
naively with a fixed three-shot demonstration template, ignores the
input and emits one of the demonstration \emph{outputs} verbatim---and
we show by direct comparison with 1.58-bit \ternary{} that this is a
property of \emph{prompting} at small-LM scale, not of 1-bit
quantization. A simple locale-conditioned rotating-demonstrations
prompting strategy resolves the qualitative cross-locale failure:
482/482 unique \bonsai{} surrogate proposals succeed (no echoes, no
errors) under this strategy, producing locale-correct,
format-preserving surrogates
(\cjk{杨娟}\,\arrowto\,\cjk{李伟}, M\"uller-Schulz\,\arrowto\,Anna
Becker, \code{11-Jul-1998}\,\arrowto\,\code{05-Aug-2003}); the fix
adds $<50$\,ms per surrogate proposal.

Substitution outperforms \code{[LABEL]}-style redaction by an enormous
margin on downstream NER training utility: at the large 400 train /
100 test scale, faker recovers F1$=$0.656\,$\pm$\,0.033 against
redact's 0.000\,$\pm$\,0.000 (a gap of many standard deviations, not
a noise effect).

Our negative finding for hybrid is the second methodological
contribution. At the matched 160/40 scale, faker (F1$=$0.506) clearly
beats hybrid (F1$=$0.346) at $p<0.001$, despite hybrid winning every
other metric (PPL, length preservation, locale-fidelity of generated
strings). The implication: \emph{natural-looking} substitution and
\emph{useful-for-training} substitution are different objectives. SLM
surrogates collapse to a narrower in-distribution surface than faker's
i.i.d.\ random outputs, and downstream NER training benefits more from
the variety than from the naturalness. We caution future work against
treating ``hybrid beats faker on PPL'' as a proxy for downstream
utility---they are not the same metric.

All code, configurations, and the 2000-document generated evaluation
corpus are released under an open license.

\section*{Reproducibility}
\label{sec:repro}

All code and the 2000-document generated evaluation set
(\code{data/samples\_2000.json}, produced by the \openaipf{}
template engine at \texttt{seed=42}) are released
at:%
\begin{center}
\url{https://github.com/asadani/on-device-pii-substitution}
\end{center}
To reproduce the primary results:

\begin{lstlisting}
# Primary metrics (Table 1, ~3 hours on CPU)
python eval_substitution.py --n 100 --modes redact faker hybrid \
       --samples-path data/samples_2000.json \
       --bonsai-size 1.7B --output results/run_v3

# Per-locale and per-template breakdowns
python analyze_results.py \
       --results results/run_v3/substitution_results.json \
       --out results/run_v3/analysis.md

# Downstream NER utility -- large scale, 3 modes (Table 2 top, ~2.5 h)
python eval_ner_utility.py \
       --samples-path data/samples_2000.json \
       --modes original redact faker --n-max 500 \
       --output results/ner_v3

# Downstream NER utility -- matched 4-mode subset (Table 2 bottom, ~1.5 h)
python eval_ner_utility.py \
       --samples-path data/samples_2000.json \
       --modes original redact faker hybrid --n-max 200 \
       --output results/ner_v3_matched

# Surrogate distinctness (Table 3, ~50 min hybrid + <1s faker on CPU)
python analyze_surrogate_distinctness.py --n-max 200 \
       --modes faker hybrid \
       --output results/surrogate_distinctness.json
\end{lstlisting}

Total wall-clock for the full pipeline is approximately 8--9 hours on
a 31\,GB single-CPU machine, dominated by hybrid-mode \bonsai{}
generation in the primary, NER, and distinctness experiments. The
privacy-filter model ($\sim$2.8\,GB) and the \xglm{} evaluator
($\sim$1.1\,GB) are downloaded once from Hugging Face on first run
and cached locally.

\section*{Use of generative AI assistance}
\label{sec:aiuse}

In line with arXiv's policy on the use of generative AI in scholarly
work, the authors disclose that a large language model assistant
(Anthropic's Claude, primarily Claude Opus 4.7) was used during the
preparation of this manuscript. Specifically: (i) the assistant
helped explore and draft prose for the introduction, related-work,
and discussion sections from the authors' notes and bullet-point
outlines; (ii) the assistant helped scaffold LaTeX, bibliography, and
table formatting; and (iii) the assistant ran the experimental
scripts and reported the resulting numerical outputs back to the
authors. All quantitative results in Tables~\ref{tab:primary},
\ref{tab:latency}, \ref{tab:ner}, and \ref{tab:distinctness} are
computed by the released code (\code{eval\_substitution.py},
\code{eval\_ner\_utility.py}, \code{analyze\_surrogate\_distinctness.py})
on the released artefact and are reproducible by any reader; no
result was generated, projected, or hallucinated by the language
model. The authors have independently reviewed all assistant-drafted
text and are solely responsible for the technical claims, methodology,
statistical analysis, and the final wording of the manuscript.

\bibliographystyle{plain}
\bibliography{refs}

\appendix
\section{Locale-conditioned demonstration pools}
\label{app:pools}

For full reproducibility we list the demonstration pools used by the
locale-conditioned rotating few-shot strategy
(Section~\ref{sec:proposers}). Per-input MD5 hashes seed a deterministic
sample of three demonstrations from the appropriate pool.

\paragraph{PERSON pool.}
\begin{itemize}[leftmargin=*, itemsep=1pt, topsep=2pt]
  \item \code{en}: (John Carter, Marcus Chen),
        (Linda Vasquez, Olivia Brennan),
        (David Kim, Theo Pemberton),
        (Sarah Patel, Maya Iyer),
        (Robert Williams, Daniel Foster),
        (Priya Krishnamurthy, Nadia Subramanian),
        (Michael O'Brien, Patrick Donovan),
        (Jennifer Wong, Cynthia Park).
  \item \code{de}: (Hans M\"uller, Karl Schmidt),
        (Anna Becker, Lena Hoffmann),
        (Klaus Wagner, Erik Kr\"uger),
        (Ingrid Weber, Petra Neumann),
        (Stefan Fischer, Dietrich Bauer),
        (Helga Zimmermann, Brigitte Klein).
  \item \code{es}: (Juan Garc\'ia, Carlos Hern\'andez),
        (Mar\'ia Rodr\'iguez, Ana Fern\'andez),
        (Diego S\'anchez, Luis Castillo),
        (Carmen Ortiz, Luc\'ia V\'azquez),
        (Roberto Jim\'enez, Pablo Morales),
        (Sof\'ia Ram\'irez, Elena Aguilar).
  \item \code{ja}: (\cjkj{山田太郎}, \cjkj{鈴木一郎}),
        (\cjkj{佐藤花子}, \cjkj{田中美咲}),
        (\cjkj{渡辺健}, \cjkj{高橋翔}),
        (\cjkj{中村裕子}, \cjkj{小林由香}),
        (\cjkj{加藤博之}, \cjkj{斎藤大輔}),
        (\cjkj{井上恵美}, \cjkj{松本香織}).
  \item \code{zh}: (\cjk{李伟}, \cjk{王芳}),
        (\cjk{张敏}, \cjk{刘洋}),
        (\cjk{陈杰}, \cjk{黄燕}),
        (\cjk{周磊}, \cjk{吴娟}),
        (\cjk{徐明}, \cjk{孙丽}),
        (\cjk{郑强}, \cjk{马晶}).
\end{itemize}

\paragraph{ADDRESS pool.} Six entries per locale; en covers both en\_US
and en\_IN. Examples include
(\code{Hauptstra\ss e 45, 10117 Berlin},
 \code{Lindenallee 12, 80331 M\"unchen}) for \code{de},
(\code{Calle Reforma 123, 06600 CDMX},
 \code{Avenida Insurgentes 456, 03100 CDMX}) for \code{es}, and
(\cjk{北京市朝阳区建国路1号}, \cjk{上海市浦东新区世纪大道100号}) for
\code{zh}. Full lists are in \code{pii\_substitute.py}.

\paragraph{DATE pool.} Pools per detected format:
\code{mdy\_slash} (5 entries), \code{ymd\_dash} (4),
\code{dmy\_dash\_mon} (4), \code{dmy\_slash} (3),
\code{unknown} (2 fallbacks).

\end{document}